\documentclass[sigconf]{acmart}

\AtBeginDocument{%
  }
\setcopyright{acmlicensed}
\copyrightyear{2025}
\acmYear{2025}
\setcopyright{cc}
\setcctype{by}
\acmConference[MM '25]{Proceedings of the 33rd ACM International Conference on Multimedia}{October 27--31, 2025}{Dublin, Ireland}
\acmBooktitle{Proceedings of the 33rd ACM International Conference on Multimedia (MM '25), October 27--31, 2025, Dublin, Ireland} 
\acmDOI{10.1145/3746027.3755765}
\acmISBN{979-8-4007-2035-2/2025/10}

\usepackage{graphicx}      
\usepackage{amsmath}        
\usepackage{booktabs}       
\usepackage{caption}        
\usepackage{algorithm}      
\usepackage{algpseudocode}  
\usepackage{array}          
\usepackage{xcolor}         
\usepackage[table,xcdraw]{xcolor} 
\usepackage[utf8]{inputenc} 
\usepackage{multirow}       
\usepackage{makecell}       
\usepackage{balance}
\newcommand{\mypara}[1]{\vspace{0.15cm}\noindent\textbf{#1}\quad}

\begin{document}

\title{CalibCLIP: Contextual Calibration of Dominant Semantics for Text-Driven Image Retrieval}


\author{Bin Kang}
\affiliation{%
  \institution{Chengdu Institute of Computer Applications, Chinese Academy of Sciences}
  \city{Chengdu}
  \country{China}}
\affiliation{%
  \institution{University of Chinese Academy of Sciences}
  \city{Beijing}
  \country{China}}
\email{kangbin23@mails.ucas.ac.cn}

\author{Bin Chen*}
\affiliation{%
  \institution{International Research Institute for Artificial Intelligence, Harbin Institute of Technology (Shenzhen)}
  \city{Shenzhen}
  \country{China}}
\email{chenbin2020@hit.edu.cn}

\author{Junjie Wang}
\affiliation{%
  \institution{Harbin Institute of Technology (Shenzhen)}
  \city{Shenzhen}
  \country{China}}
\email{junjiewang@stu.hit.edu.cn}

\author{Yulin Li}
\affiliation{%
  \institution{Harbin Institute of Technology (Shenzhen)}
  \city{Shenzhen}
  \country{China}}
\email{yulinli@stu.hit.edu.cn}

\author{Junzhi Zhao}
\affiliation{%
  \institution{Southwest Jiaotong University}
  \city{Chengdu}
  \country{China}}
\email{zhaojunzhi@my.swjtu.edu.cn}

\author{Junle Wang}
\affiliation{%
  \institution{Tencent}
  \city{Shenzhen}
  \country{China}}
\email{jljunlewang@tencent.com}

\author{Zhuotao Tian}
\authornote{Corresponding author.}
\affiliation{%
  \institution{Harbin Institute of Technology (Shenzhen)}
  \city{Shenzhen}
  \country{China}}
\email{tianzhuotao@gmail.com}

\renewcommand{\shortauthors}{Bin Kang et al.}

\begin{abstract}
  Existing Visual Language Models (VLMs) suffer structural limitations where a few low contribution tokens may excessively capture global semantics, dominating the information aggregation process and suppressing the discriminative features in text-driven image retrieval tasks. To address this, we introduce \textbf{CalibCLIP}, a training-free method designed to calibrate the suppressive effect of dominant tokens. Specifically, in the visual space, we propose the Contrastive Visual Enhancer (CVE), which decouples visual features into target and low information regions. Subsequently, it identifies dominant tokens and dynamically suppresses their representations. 
  \begin{figure}[htp]
    \centering
    \includegraphics[width=1.0\linewidth]{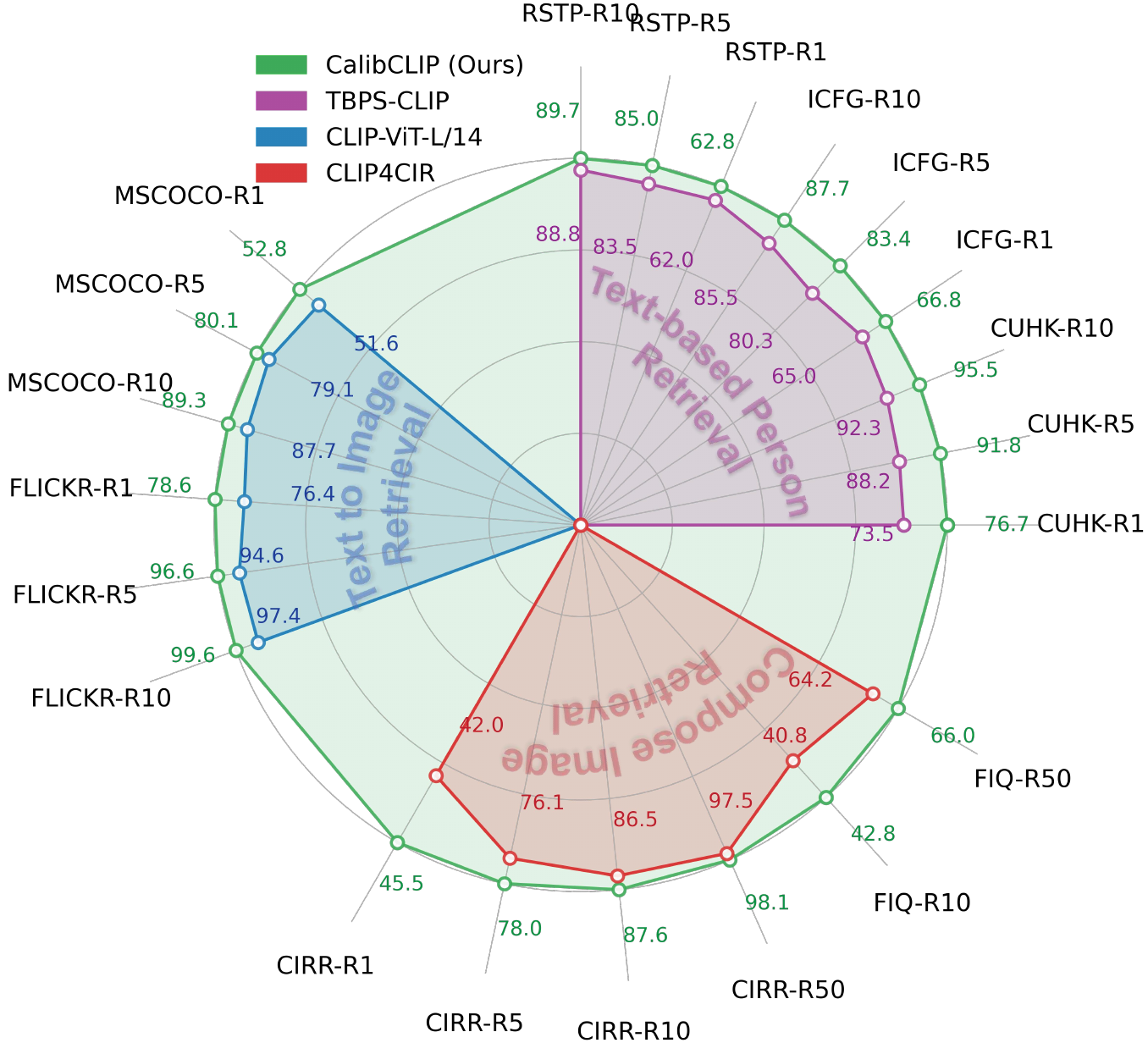} 
    \caption{CalibCLIP has achieved significant performance gains across multiple test benchmarks in three text-driven image retrieval paradigms.}
    \label{fig:performance}
\end{figure}
  In the textual space, we introduce the Discriminative Concept Calibrator (DCC), which aims to differentiate between general and discriminative concepts within the text query. By mitigating the challenges posed by generic concepts and improving the representations of discriminative concepts, DCC strengthens the differentiation among similar samples. Finally, extensive experiments demonstrate consistent improvements across seven benchmarks spanning three image retrieval tasks, underscoring the effectiveness of CalibCLIP. Code is available at: \url{https://github.com/kangbin98/CalibCLIP}
\end{abstract}




\maketitle

\section{Introduction}
Recently, Vision-Language Models (VLMs) \cite{10445007, NEURIPS2021_50525975, Zhai_2023_ICCV, Alayrac2022Flamingo, Lin_2024_CVPR} built upon Vision Transformer (ViT) \cite{dosovitskiy_2021_vit} architectures have achieved remarkable advancements across multiple domains, establishing a solid foundation for Text-Driven Image Retrieval (TDIR) tasks \cite{Liu_2021_ICCV, TPAMI_2024, Ray2023Cola, Yang2022Dual, luo2025imagescope}. Current approaches \cite{Cheng_2022, NEURIPS2024_fe759454} leverage multimodal representations through cross-modal aligned features extracted from VLMs, achieving robust open-domain generalization.

\begin{figure}[!t]
\centering
\includegraphics[width=.95\linewidth, height=5.5cm]{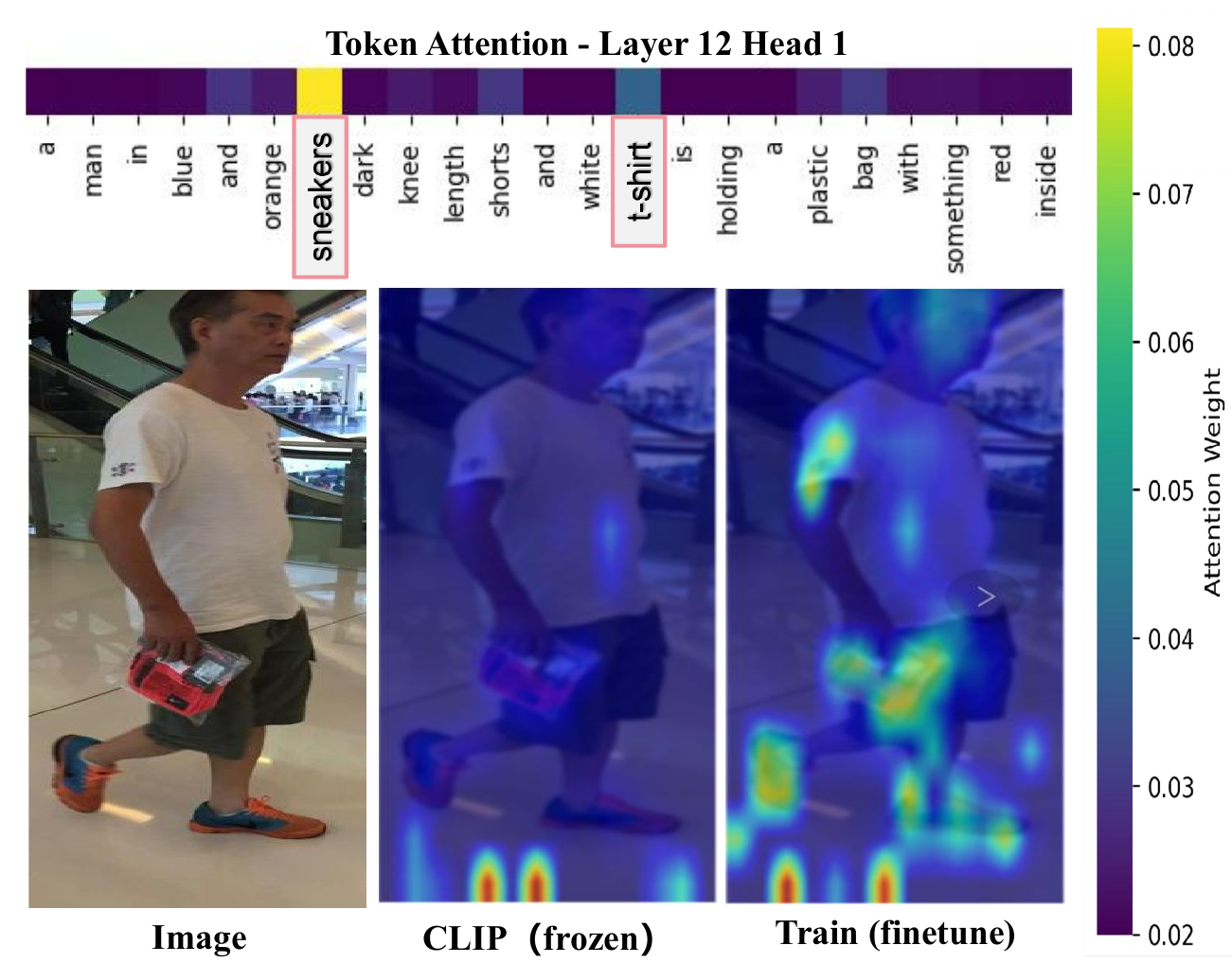} 
\caption{CLS/EOT token self-attention. A few low information tokens receive disproportionately high attention, persisting even with task-specific fine-tuning.}
\label{fig:cls/EOT}
\end{figure}

However, VLM-based retrieval methods predominantly depends on \textit{global semantic token alignment} mechanisms, where cross modal matching is achieved through the similarity between aggregated global semantic tokens. This approach presents a significant bottleneck for establishing nuanced associations. Some studies \cite{Yang_TCSVT, Liu_2021_ICCV} establish fine-grained cross-modal correspondences through patch-level interactions, but these methods are susceptible to noisy tokens, such as local image patches that belong to different objects but share identical appearances. Mainstream approaches \cite{Jiang_2023_CVPR, ACMMM_2024_Li} employ self-attention mechanisms to interact with patch tokens and aggregate information, generating a single global semantic token (e.g., [CLS] and [EOT] tokens) for cross-modal matching. Nevertheless, due to the absence of explicit supervisory guidance during the information aggregation, a critical question arises: \textit{Can the existing aggregation process effectively focus on the discriminative tokens when constructing the global proxy for cross-modal alignment?}

\mypara{Key Observations. }
To explore this, inspired by \cite{DBLP}, we analyze the feature attention activation states of widely used vision and text encoders \cite{pmlr-v139-radford21a} in TDIR tasks. As illustrated in Figure~\ref{fig:cls/EOT}, the image and text semantic spaces exhibit significant attention bias, with only a few tokens receiving high attention. These tokens carry excessive global semantics and dominate information propagation during self-attention. In the visual space, the semantic dominant tokens are gathered in low information regions, such as the background, forming spatially invariant outlier tokens of the target object. In the text space, the semantic dominance phenomenon is primarily observed as over-reliance on common attributes, consequently, hindering the effective representation of discriminative features. 
\textit{These contextually dominant tokens may impede the model's capacity to prioritize discriminative features, resulting in a diminished focus on visual local information and textual discriminative concepts, making it hard to differentiate highly similar samples.}

\begin{figure}[!t]
\centering
\includegraphics[width=.95\linewidth, height=5.5cm]{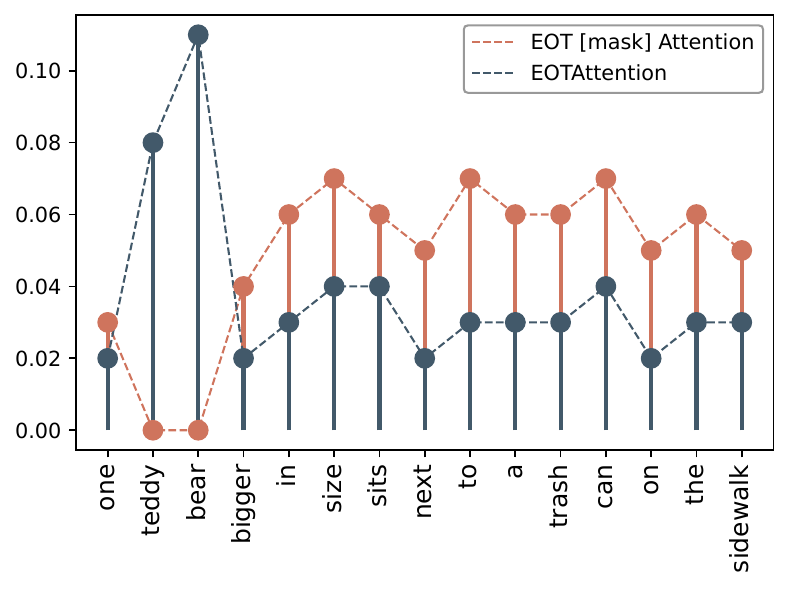} 
\caption{Comparison of [EOT] token attention: When dominant words like ``teddy"  and ``bear" are masked, attention on the remaining tokens significantly increases.}
\label{fig:EOT token}
\end{figure}

\mypara{Our Solution.}
To tackle the aforementioned challenges, we propose CalibCLIP, a training-free approach aimed at alleviating issues stemming from contextually dominant tokens. In the visual domain, we introduce the Contrastive Visual Enhancer (CVE) to separate visual features into target regions and low information regions. We then employ a dynamic approach to identify and suppress the dominant tokens, thereby improving the representations of local visual details. In the textual domain, we design a Discriminative Concept Calibrator (DCC) that disentangles text into general and discriminative attributes. By suppressing the influence of general attributes and emphasizing that of discriminative attributes, DCC substantially enhances the model’s ability to distinguish between semantically similar concepts.

\begin{figure*}[!ht]
\centering
\includegraphics[width=.99\textwidth]{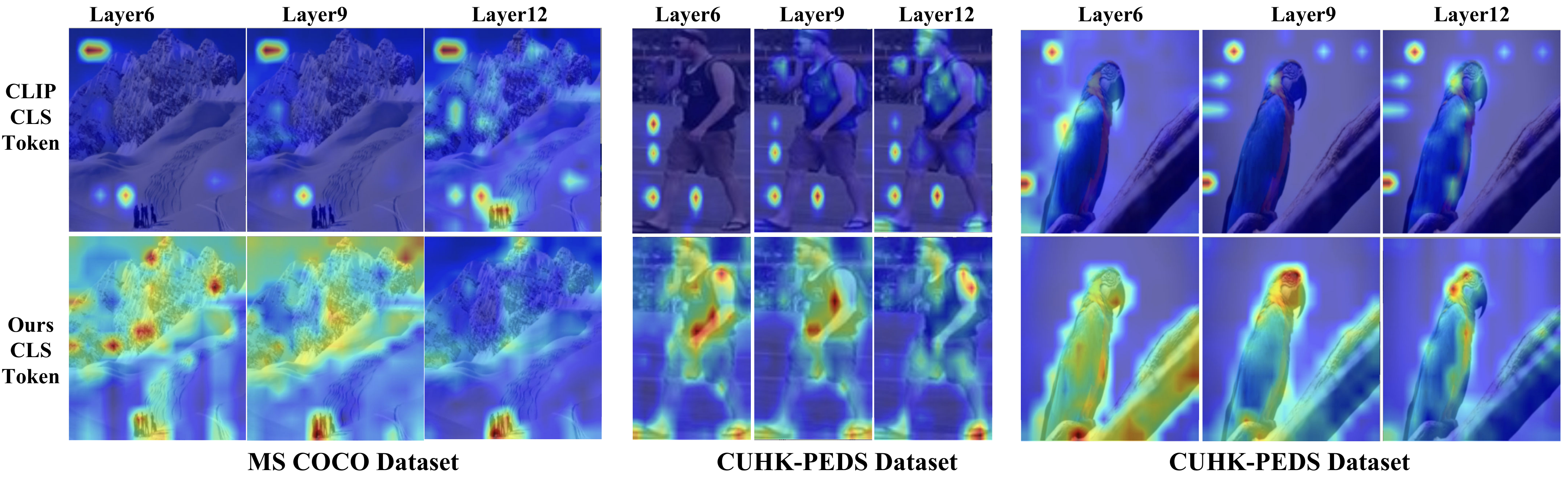} 
\caption{Visualizing attention maps across encoding layers shows the baseline model's tendency to over-focus on low information tokens, whereas our method prioritizes task-relevant regions.} 
\label{fig:cls}
\end{figure*}

To validate the effectiveness of CalibCLIP, we conduct comprehensive evaluations across seven standardized benchmarks covering three retrieval paradigms, including Text-based Person Retrieval (TBPR), Text-to-Image Retrieval (TIR), and Compose Image Retrieval (CIR). As illustrated in Figure~\ref{fig:performance}, compared to the baseline model, CalibCLIP  achieved improvements of 2.27\%, 1.70\%, and 1.96\%  in Rank@K performance for these tasks, respectively, without the need for additional training. In summary, our contributions are as follows:

\begin{itemize}
    \item Our study presents the key issues that could impede the efficacy of cross-modal retrieval: attention weights are misdirected during global information aggregation, shifting focus from genuinely informative tokens to low information, contextually dominant tokens.
    \item  To tackle this challenge, we introduce CalibCLIP, a training-free approach that combines Contrastive Visual Enhancer (CVE) and Discriminative Concept Calibrator (DCC) to alleviate the impact of contextually dominant tokens, thereby improving detailed visual features and distinctive text signals.
    \item CalibCLIP demonstrates consistent effectiveness and versatility across different scenarios. Extensive evaluations on seven benchmarks confirm its robust performance and generalizability to various Text-to-Image Retrieval (T2IR) architectures.
\end{itemize}

\section{Related Work}
\label{sec:related_work}
\mypara{Visual Language Models.} VLMs\cite{Lai_2024_CVPR, Chen_2024_CVPR, Zhai_2023_ICCV, Guan_2024_CVPR} such as CLIP excel in achieving cross-modal alignment through contrastive pre-training but encounter a structural limitation where scattered attention hinders nuanced feature discrimination. Recent studies \cite{NEURIPS2024_a0303731, Zhang_2024_CVPR} have highlighted disruptive tokens and proposed various solutions for fine-grained tasks like open-vocabulary segmentation \cite{Shao-eccv-2024} and object detection\cite{Hamilton_2024_CVPR}. However, these methods often require additional training and currently lack a unified and comprehensive analysis and solution for image retrieval. In response, we introduce a straightforward, training-free, dual-space calibration framework that suppresses dominant token representations without the need for extra training, thereby enhancing fine-grained perception in retrieval tasks.

\mypara{Text-driven image retrieval.} Building on the achievements of large language models (LLMs) \cite{longlora, bai2023qwentechnicalreport}, the field of text-driven image retrieval \cite{Xu2024Invisible, Liao2024Selection, liu2024candidate, bai_2023_sentence, APTM_MM_2023, Feng_ACMMM_2024}has made significant strides. However, existing methods \cite{Wan_2024_CVPR, Fu_2023_CVPR} mainly focus on global alignment between images and text, often neglecting fine-grained details and struggling with intricate queries. This issue is particularly pronounced in text-based person retrieval tasks \cite{tcsvt_2023_CTLG,tnnls_2024_EAIB, Liu2024Causality}, which require precise modeling of subtle attributes and spatial relationships. Yet, current models primarily concentrate on associations at the object level. While recent compositional retrieval methods \cite{Pan_2023_CVPR, 2024_SIGIR_ACM} extend semantic alignment to multi-concept queries, they are hindered by high computational demands and limitations in data scalability. This hampers both fine-grained perception and the performance of compositional queries. To address these challenges, we propose a training-free approach that tackles the semantic dominance of abnormal tokens in the shared embedding space, thereby enhancing fine-grained perception and cross-modal alignment.

\begin{figure*}[!ht]
\centering
\includegraphics[width=.99\textwidth, height=10cm]{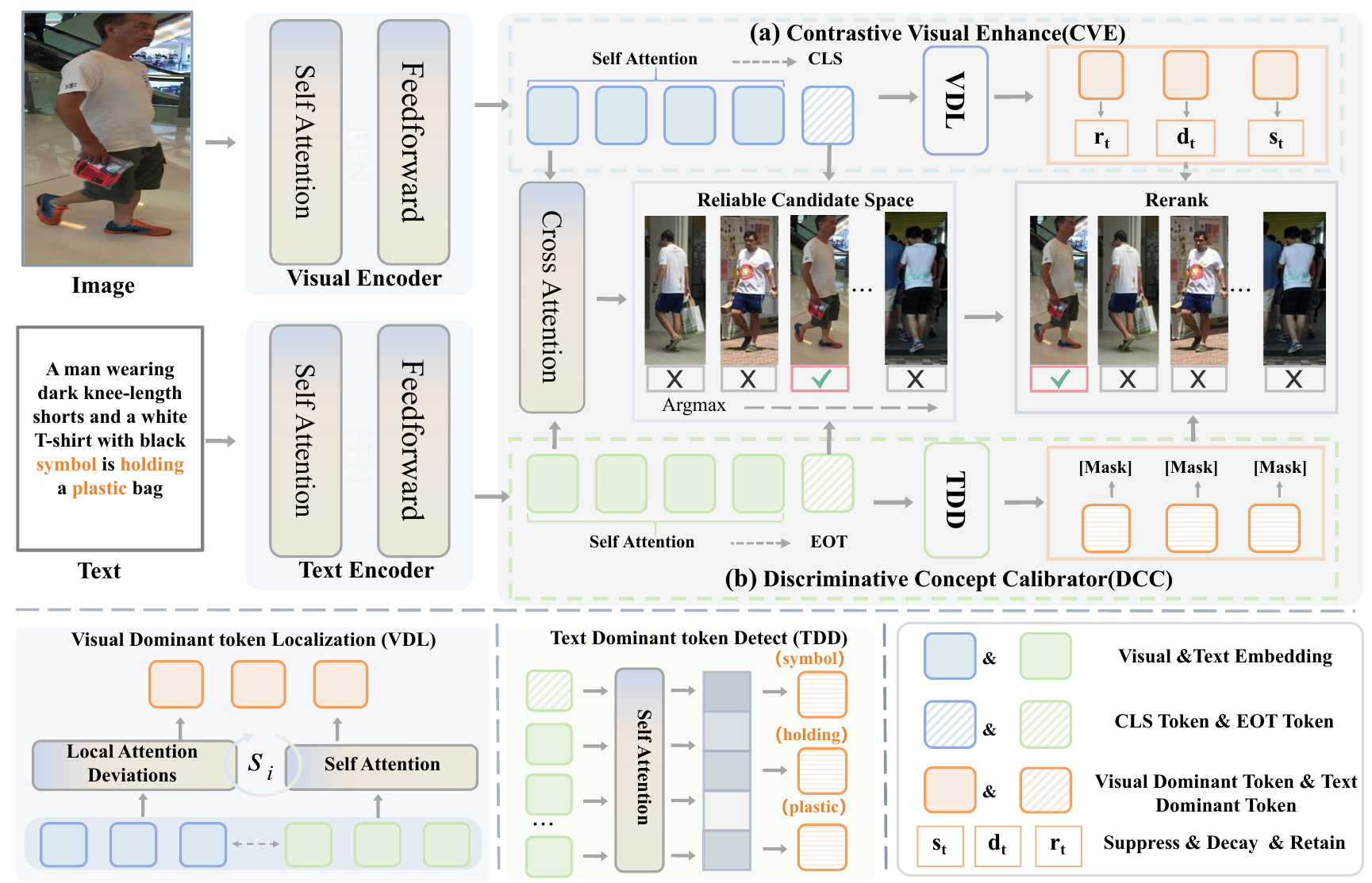} 
\caption{Illustration of CalibCLIP framework. We calibrate contextually dominant tokens through a dual-space intervention: In visual space, the CVE module isolates objects from low information regions while suppressing dominant tokens. In text space, the DCC module disentangles text into general and discriminative attributes for fine-grained differentiation. } 
\label{fig:overview}
\end{figure*}

\section{Preliminary}
\label{sec:preliminary}

\mypara{VLM-based image retrieval architecture.} 
The VLM-based image retrieval architecture \cite{ACMMM_2024, Gu_2024_CVPR, NEURIPS2024_d4ee9e80} typically adopts a dual encoder framework similar to CLIP, which encodes the input image \(\mathbf{I}_i\) and text \(\mathbf{T}_i\) into a series of visual features \(\{\mathbf{v}_{\text{cls}}, \mathbf{v}_1, \dots, \mathbf{v}_n\}\) and text features \(\{\mathbf{t}_{\text{eot}}, \mathbf{t}_1, \dots, \mathbf{t}_m\}\). Here, \(\mathbf{v}_{\text{cls}}\) and \(\mathbf{t}_{\text{eot}}\) denote the global representations of the image and text query, obtained by aggregating local features via attention mechanism:

\begin{equation}
\mathbf{v}_{\text{cls}} = \sum_{i=0}^{N} \mathop{\mathrm{softmax}}\left(\frac{\mathbf{Q}_{\text{cls}} \mathbf{K}_i^\top}{\sqrt{d}}\right)  \mathbf{V}_i,
\end{equation}
where \(\mathbf{Q}_{\text{cls}}\) represents the [CLS] token query, and \(\mathbf{K}\) and \(\mathbf{V}\) denote the key and value of the \(i\)-th patch token, with d denoting the dimension of each attention head, \(\mathbf{t}_{\text{eot}}\) is computed analogously.

\mypara{Motivation.}
We observe that a few tokens may dominate global semantic aggregation, primarily occupying low information regions or general attributes. These contextually dominant tokens could potentially hinder the representations of subtle cues that are essential for distinguishing various identities in the retrieval process. To assess the impacts, we conducted the following experiments.

\textit{In the visual domain}, we visualize attention patterns between the [CLS] token and patch tokens within the visual encoder's last layer. As illustrated in Figure~\ref{fig:cls}, the [CLS] token allocates excessive attention to a few low information patches, thereby suppressing its focus on the target regions. Therefore, we posit that suppressing high-attention tokens in low information regions can redirect the [CLS] token's focus toward target regions.  

\textit{In the textual domain}, we similarly obtain attention distributions between the [EOT] token and subword units. As shown in Figure~\ref{fig:EOT token}, the [EOT] token exhibits an over-reliance on generic attribute tokens, which diminishes its sensitivity to more discriminative textual cues. Therefore, we hypothesize that alleviating the contextual overwhelming effect caused by general attributes leads to significantly enhanced attention activation in the remaining tokens.

These findings underscore the influence of contextually dominant tokens on model performance, emphasizing the significance of our study. 

\section{Method}
In this section, we first introduce the model architecture in Section~\ref{sec:overview}. After that, Section~\ref{sec:contrastive_visual_enhancer} presents our visual dominant token calibration method, which reduces the obstruction of target region information. Section~\ref{sec:discriminative_concept_calibrator} details the text-dominant token calibration approach for alleviating suppression of discriminative attribute tokens.

\subsection{Overview}
\label{sec:overview}
As illustrated in Figure~\ref{fig:overview}, to address the degradation in fine-grained perception capabilities caused by the dominant tokens during cross-modal retrieval, we propose CalibCLIP, a training-free framework that mitigates the contextual dominant issues via calibrations in both visual and textual spaces.

Specifically, in the visual space, the framework establishes the Contrastive Visual Enhancer (CVE), as shown in Figure~\ref{fig:overview} (a), which separates the target regions from regions with low information content. Subsequently, it identifies and localizes visual dominant tokens, and dynamically reduces their impact to enhance visual feature representation. 
In parallel, within the textual space, we introduce a Discriminative Concept Calibrator (DCC) (Figure~\ref{fig:overview} (b)) to disentangle generic and discriminative concepts within the text query. This process suppresses the overwhelming influence of generic attribute tokens and enhances the representation of discriminative cues, thereby benefiting the distinction between similar concepts.


\subsection{Contrastive Visual Enhancer}
\label{sec:contrastive_visual_enhancer}
To alleviate interference from visual dominant tokens, we introduce the Contrastive Visual Enhancer (CVE), which identifies and reduces the activations of such tokens in regions with low information content to uncover more visual details. The proposed CVE comprises three essential steps, as outlined below.

\mypara{Step-1: Visual space decoupling.}
To mitigate the impact of visual dominant tokens in low-information regions, we decouple visual features by leveraging semantic correlations between image patches and textual descriptions. 
Specifically, we compute the cosine similarity between each image patch token $\mathbf{v}_i \in \mathbb{R}^{N \times d}$ (where $N$ denotes the number of patches) obtained from CLIP’s visual encoder and the [EOT] token $\mathbf{t}_{\text{eot}} \in \mathbb{R}^d$ that captures the global textual semantics $S(\mathbf{v}_i, \mathbf{t}_{\text{eot}})$.
Having observed that target regions typically demonstrate greater similarity to the text query compared to low-information regions, we introduce an adaptive threshold $\tau = \frac{1}{N} \sum_{i=1}^N S(\mathbf{v}_i, \mathbf{t}_{\text{eot}})$, generating a binary mask $M_g(\mathbf{v}_i)$ where $M_g=1$ if $S(\mathbf{v}_i, \mathbf{t}_{\text{eot}}) \geq \tau$, and $M_g=0$ otherwise. Further analysis is detailed in Section~\ref{sec:Visual_Feature_Decoupling_Threshold}. Here, $M_g=1$ represents the high-correlation region with the text, referred to as the target region $V_{\text{f}}$, and therefore $M_g=0$ corresponds to the low-information background regions $V_{\text{b}}$:
\begin{equation}
V_{\text{f}} = M_g \cdot \mathbf{v}_i, \quad V_{\text{b}} = (1 - M_g) \cdot \mathbf{v}_i.
\end{equation}


\mypara{Step-2: Dominant Token Localization.} 
As noted in Section~\ref{sec:preliminary}, visual dominant tokens are often found in the low-information background regions. To address the adverse effects caused by these tokens, we introduce a method in this step to detect visual dominant tokens by simultaneously evaluating self-attention scores and local attention deviations in low-information regions. 

Specifically, each visual token $\mathbf{v}_i$ is first assessed by its self-atten-tion score $\mathbf{A}(\mathbf{v}_i)$ relative to the [CLS] token. As the self-attention score $\mathbf{A}(\mathbf{v}_i)$ reflects the global relevance, tokens with high $\mathbf{A}(\mathbf{v}_i)$ values are prioritized as potential candidates for visual dominant tokens. 

The attention score $\mathbf{A}(\mathbf{v}_i)$ evaluates the overall relevance, yet dominant tokens also demonstrate significant semantic deviations within their local contexts, as illustrated in Figure~\ref{fig:cls}.
Hence, we introduce the local attention deviations $LC({\mathbf{v}_i})$, which quantifies attention deviations with respect to local neighbors.
For each token $\mathbf{v}_i$, we designate its 8-connected spatial neighbors as $\mathcal{N}_i = {\mathbf{v}_j \mid |\text{p}_i - \text{p}_j|_2 \leq 1}$, which includes the neighboring patches. Then, the local attention deviations $LC({\mathbf{v}_i})$ is calculated as:
\begin{equation}
LC({\mathbf{v}_i}) = \frac{{\mathbf{A}({\mathbf{v}_i}) - \frac{1}{{\left| {{N_i}} \right|}}\sum\nolimits_{j \in {N_i}} {\mathbf{A}({\mathbf{v}_j})} }}{{\sqrt {\frac{1}{{\left| {{N_i}} \right|}}\sum\nolimits_{j \in {N_i}} {{{(\mathbf{A}({\mathbf{v}_j}) - {\mu _{{N_i}}})}^2}} }  + \varepsilon }}
\end{equation}
where $\frac{1}{|\mathcal{N}_i|}\sum_{j \in \mathcal{N}_i} \mathbf{A}(\mathbf{v}_j)$ represents the mean attention score of the neighborhood, $\sqrt {\frac{1}{{\left| {{N_i}} \right|}}\sum\nolimits_{j \in {N_i}} {{{(\mathbf{A}({\mathbf{v}_j})}}}}$ represents the standard deviation, and $\mu _{N_i}$ denotes the mean of neighbourhood  token attention. In essence, the local attention deviations $LC({\mathbf{v}_i})$ gauge the disparity in attention between visual tokens and their local context, which can be leveraged to identify the visual dominant tokens.

Consequently, tokens exceeding the local attention deviation values of all their neighboring tokens are selected as the visual dominant tokens.

\mypara{Step-3: Context-adaptive feature rectification.}  
After identifying the visual dominant tokens, an intuitive way is to remove them directly. However, direct token removal risks disrupting spatial coherence in feature maps, leading to inferior performance. 
To address this issue, we introduce a context-adaptive rectification mechanism that preserves structural integrity while modulating visual dominant features. 

The rectification process operates on the combined score $s = LC(\mathbf{v}_i) \cdot \mathbf{A}(\mathbf{v}_i)$, which integrates local contextual discrepancies and global relevance. Subsequently, we can define the feature rectifier $g(s)$ based on $s$:
\begin{equation}
g(s) = \begin{cases} 
1 & s < \tau \\
\eta & s \geq \tau 
\end{cases}
\end{equation}
In this context, the rectifier $g(s)$ aims at adaptively adjusting feature magnitudes: preserving original representations for semantically aligned tokens ($s < \tau$) and constraining over-expressive features to a residual level $ \eta $ ($s \geq \tau$).

Finally, the rectified features can be obtained by $\hat{\mathbf{v}}_i = \mathbf{v}_i \cdot g(s)$ so that the spatial coherence can be maintained. This process enhances cross-modal alignment by redistributing attention toward textually relevant regions without compromising structural integrity, thereby improving the model's ability to reconcile visual and linguistic patterns.

\subsection{Discriminative Concept Calibrator}
\label{sec:discriminative_concept_calibrator}
Section~\ref{sec:contrastive_visual_enhancer} addresses the issues caused by visual dominant tokens. However, our findings in Section~\ref{sec:preliminary} show that, in the textual domain, tokens belonging to general concepts may overwhelm the representations of the discriminative concepts, causing difficulties for cross-modal retrieval. Therefore, in this section, we introduce the Discriminative Concept Calibrator (DCC) to improve the representation of subtle differences among different samples by effectively mitigating the influences of generic concepts while maintaining performance. The proposed DCC comprises three steps outlined below.

\begin{table*}[ht]
\centering
\caption{Performance comparison of TBPR on CUHK-PEDES, ICFG-PEDES and RSTPReid test sets(w/o CLIP: methods without CLIP backbone; w/ CLIP: methods with CLIP backbone)}
\label{tab:TBPRresults} 
\begin{tabular*}{\linewidth}{@{\extracolsep{\fill}} l|c|cccc|cccc|cccc}
\hline
 & \multirow{2}{*}{\textbf{Methods}} & \multicolumn{4}{c|}{\textbf{CUHK-PEDES}} & \multicolumn{4}{c|}{\textbf{ICFG-PEDES}} & \multicolumn{4}{c}{\textbf{RSTPReid}} \\
\cline{3-14}
& & \textbf{R@1} & \textbf{R@5} & \textbf{R@10} & \textbf{mAP} & \textbf{R@1} & \textbf{R@5} & \textbf{R@10} & \textbf{mAP} & \textbf{R@1} & \textbf{R@5} & \textbf{R@10} & \textbf{mAP} \\ 
\hline

\multirow{4}{*}{\rotatebox[origin=c]{90}{w/o CLIP}} 
& EAIBC \cite{tnnls_2024_EAIB} & 64.96 & 83.36 & 88.42 & -- & 58.95 & 75.95 & 81.72 & -- & 49.85 & 70.15 & 79.85 & -- \\
& IVT \cite{ssan_ECCV} & 65.59 & 83.11 & 89.20 & -- & 56.04 & 73.60 & 80.22 & -- & 46.70 & 70.00 & 78.80 & -- \\
& CTLG \cite{tcsvt_2023_CTLG} & 69.47 & 87.13 & 92.13 & 60.56 & 57.69 & 75.79 & 82.67 & -- & -- & -- & -- & -- \\
& SAP-SAM \cite{ACMMM_2024_SAM} & 75.05 & 89.93 & 93.73 & -- & 63.97 & 80.84 & 86.17 & -- & 62.85 & 82.65 & 89.85 & -- \\
\hline
\hline

\multirow{9}{*}{\rotatebox[origin=c]{90}{w/ CLIP}}
& CFine\cite{TIP_2023} & 69.57 & 85.93 & 91.15 & -- & 60.83 & 76.55 & 82.42 & -- & 50.55 & 72.50 & 81.60 & -- \\ 
& IRRA\cite{Jiang_2023_CVPR} & 73.38 & 89.93 & 93.71 & 66.13 & 63.46 & 80.25 & 85.82 & 38.06 & 60.20 & 81.30 & 88.20 & 47.17 \\ 
& TILT\cite{ACMMM_2024} & 74.46 & 90.21 & 94.19 & 66.31 & 63.77 & 80.80 & 86.00 & 38.07 & 60.75 & 81.80 & 88.70 & 47.56 \\ 
& IRLT\cite{Liu_Qin_Chen_Cheng_Yang_2024} & 74.46 & 90.19 & 94.01 & -- & 64.72 & 81.35 & 86.31 & -- & 61.49 & 82.26 & 89.23 & -- \\
\cline{2-14}
& CLIP-ViT/16 & 66.54 & 86.94 & 91.77 & 62.69 & 57.44 & 75.79 & 82.22 & 33.03 & 56.67 & 78.09 & 86.62 & 44.25 \\ 
& \cellcolor{gray!10}\hspace{1em}\textbf{+ CalibCLIP} & \cellcolor{gray!10}\textbf{71.88} & \cellcolor{gray!10}\textbf{90.50} & \cellcolor{gray!10}\textbf{94.75} & \cellcolor{gray!10}\textbf{65.22} & \cellcolor{gray!10}\textbf{62.54} & \cellcolor{gray!10}\textbf{80.18} & \cellcolor{gray!10}\textbf{84.57} & \cellcolor{gray!10}\textbf{37.37} & \cellcolor{gray!10}\textbf{60.30} & \cellcolor{gray!10}\textbf{82.78} & \cellcolor{gray!10}\textbf{88.66} & \cellcolor{gray!10}\textbf{46.47} \\
& TBPS-CLIP\cite{Cao_Bai_Zeng_Ye_Zhang_2024} & 73.54 & 88.19 & 92.35 & 65.38 & 65.05 & 80.34 & 85.47 & 39.83 & 61.95 & 83.55 & 88.75 & 48.26 \\ 
& \cellcolor{gray!10}\hspace{1em}\textbf{+ CalibCLIP} & \cellcolor{gray!10}\textbf{76.72} & \cellcolor{gray!10}\textbf{91.80} & \cellcolor{gray!10}\textbf{95.47} & \cellcolor{gray!10}\textbf{67.58} & \cellcolor{gray!10}\textbf{66.78} & \cellcolor{gray!10}\textbf{83.40} & \cellcolor{gray!10}\textbf{87.73} & \cellcolor{gray!10}\textbf{41.81} & \cellcolor{gray!10}\textbf{62.82} & \cellcolor{gray!10}\textbf{85.02} & \cellcolor{gray!10}\textbf{89.71} & \cellcolor{gray!10}\textbf{50.51} \\
\hline
\end{tabular*}
\label{integrated_table}
\end{table*}

\mypara{Step-1: Textual subspace decomposition.}
To address the over-reliance of the [EOT] token on generic semantic attributes, we start by disentangling the text representations into two complementary subspaces. 

Specifically, for an \( L \)-layer text encoder, let \(\mathbf{A}^l \in \mathbb{R}^{1 \times n}\) denote the attention weights between the [EOT] token and \( n \) subword tokens at layer \( l \). We compute layer-wise importance \(\gamma_l = \|\mathbf{h}^l_{\text{eot}}\|_2\) to weight each layer's contribution. \(\gamma_l\) quantifies the activation magnitude of the [EOT] token's hidden state $\mathbf{h}^l_{\text{eot}}$, and it has been observed that the feature magnitude is directly related to its importance \cite{NEURIPS2022_2ce10f14}. Therefore, the aggregated attention for the $i$-th token is computed as:  
\begin{equation}
\alpha_i = \frac{\sum_{l=1}^L \gamma_l \mathbf{A}_i^l}{\sum_{l=1}^L \gamma_l}
\end{equation}
$\alpha_i$ quantifies the total attention allocated to the $i$-th token across all encoder layers.  

Subsequently, given $\alpha_i$, we can decompose textual features into two complementary subspaces with the threshold $\tau_t$: 
\begin{itemize}
    \item \textbf{General Attribute Subspace:} Tokens with high attention values (\( \alpha_j \geq \tau_t \)) contribute to the representation of generic concepts \(\mathbf{t}_g \in \mathbb{R}^d\), encoding coarse-grained attributes (e.g., object categories like ``apparel" or ``animal") that facilitate cross-modal alignment due to their strong correlation with visual primitives.
    \item\textbf{Discriminative Attribute Subspace:} Tokens with low attention values (\(\alpha_k < \tau_t \)) form the features of discriminative concepts \(\mathbf{t}_d \in \mathbb{R}^d\), capturing more detailed characteristics (e.g., ``striped texture" or ``running action") essential for precise visual comprehension but often overlooked during the attention process.

\end{itemize}

\mypara{Step-2: Adaptive semantic modulation.}
With the decoupled subspaces for general and discriminative attributes respectively, we can alleviate the over-reliance on the general attributes with an adaptive token masking strategy, which can be formulated as:
\begin{equation}  
\label{eq:token_masking}
t_{\text{a}} = \sum_{i=1}^n \left[ (1 - m_i) \cdot \mathbf{t}_g \right]  
\end{equation}  
In Eq.~\eqref{eq:token_masking}, the features of the general attribute tokens \(\mathbf{t}_g\) are scaled by a modulation coefficient \(m_i\in[0,1]\), yielding the attenuated representation \(\mathbf{t}_a\in\mathbb{R}^d\).  We define the coefficient as  $m_i = \frac{|\mathcal{D}|}{|\mathcal{G}+\mathcal{D}|}$, where \(|\mathcal{G}|\) and\(|\mathcal{D}|\) is the number of general and discriminative attributes.  
Intuitively, when text descriptions contain abundant discriminative details (i.e., \(|\mathcal{D}| \gg |\mathcal{G}|\)), the mask values tend toward \(m_i \to 1\), suppressing \(\mathbf{t}_g\) while enhancing \(\mathbf{t}_d\) to prioritize fine-grained distinctions.  
Conversely, when discriminative cues are scarce (i.e., \(|\mathcal{G}| \gg |\mathcal{D}|\)), the mask values decay toward \(m_i \to 0\), preserving \(\mathbf{t}_g\) to maintain semantic stability and ensure robust cross-modal alignment under sparse textual descriptions.

\mypara{Step-3: Inference with discriminative similarity.}
The cross-modal retrieval result is typically predicted by selecting the sample pair with maximal similarity $\text{sim}(\mathbf{t}_{\text{eot}}, \mathbf{v}_{\text{cls}})$ between the text's terminal [EOT] token $\mathbf{t}_{\text{eot}}$ and the image's [CLS] token $\mathbf{v}_{\text{cls}}$~\cite{NEURIPS2024_8b54ecd9, Koley_2024_CVPR}. 
While the modulated feature $\mathbf{t}_a$ (from Step-2) alleviates contextual dominance bias, the conventional similarity measurement $\text{sim}(\mathbf{t}_{\text{eot}}, \mathbf{v}_{\text{cls}})$ remains suboptimal for capturing fine-grained discrepancies. This limitation arises because the [EOT] token inherently encapsulates global textual semantics, potentially obscuring nuanced discriminative features that reside in intermediate token interactions.

To mitigate this limitation, we establish a new token with enhanced discriminative cues to complement the [EOT] token. First, we introduce a new token $\mathbf{r}$, concatenated with the modulated feature $\mathbf{t}_a$ and discriminative feature $\mathbf{t}_d$, to be processed through a self-attention layer:
\begin{equation}
   \mathbf{t}_{\text{r}},\_ ,\_ = \text{Self-Attention}([\mathbf{r}; \mathbf{t}_a; \mathbf{t}_d])
\end{equation}  
$\mathbf{t}_{\text{r}}$ denotes the self-attention output of the learnable token $\mathbf{r}$. Then, a cross-attention layer is adopted to generate $\mathbf{\hat{t}}_{\text{r}}$, with $\mathbf{t}_{\text{r}}$ as the query, and the key and value being the concatenation of $\mathbf{t}_a$ and $\mathbf{t}_d$.
In this context, $\mathbf{\hat{t}}_{\text{r}}$ incorporates essential fine-grained information from the modulated feature $\mathbf{t}_a$ and discriminative feature $\mathbf{t}_d$ to enhance the differentiation between similar concepts.

\begin{table*}[ht]
\centering
\caption{Performance comparison of TIR on Flickr30K and MSCOCO test sets.}
\label{tab:TIRresults}
\begin{tabular*}{\linewidth}{@{\extracolsep{\fill}} c|l|rrr|rrr|rrr|rrr}
\hline
\multirow{3}{*}{\rotatebox{90}{\textbf{}}} & \multirow{3}{*}{\textbf{Method}} & \multicolumn{6}{c|}{\textbf{MSCOCO(5K Text Set)}} & \multicolumn{6}{c}{\textbf{Flickr30K(1K Text Set)}} \\ 
\cline{3-14}
& & \multicolumn{3}{c|}{\textbf{Text to Image}} & \multicolumn{3}{c|}{\textbf{Image to text}} & \multicolumn{3}{c|}{\textbf{Text to Image}} & \multicolumn{3}{c}{\textbf{Image to text}} \\
\cline{3-5} \cline{6-8} \cline{9-11} \cline{12-14}
& & \textbf{R@1} & \textbf{R@5} & \textbf{R@10} & \textbf{R@1} & \textbf{R@5} & \textbf{R@10} & \textbf{R@1} & \textbf{R@5} & \textbf{R@10} & \textbf{R@1} & \textbf{R@5} & \textbf{R@10} \\

\hline\hline
\multirow{8}{*}{\rotatebox{90}{w/o CLIP}}  
& VSE$\infty$ \cite{Chen_2021_CVPR} 
& 39.30 & 69.90 & 81.10 & 56.60 & 83.60 & 91.40 & 56.40 & 83.40 & 89.90 & 76.50 & 94.20 & 97.70 \\
& NAAF \cite{Zhang_2022_CVPR} 
& 42.50 & 70.90 & 81.40 & 58.90 & 85.20 & 92.00 & 61.00 & 85.30 & 90.60 & 81.90 & 96.10 & 98.30 \\
& HREM \cite{Fu_2023_CVPR}
& 41.30 & 71.90 & 82.40 & 60.60 & 86.40 & 92.50 & 60.90 & 85.60 & 91.30 & 81.40 & 96.50 & 98.50 \\
& NUIF \cite{Zhang_Zhang_Zhang_Mao_2024} 
& 43.30 & 72.40 & 82.60 & 61.80 & 86.80 & 93.10 & 60.70 & 85.00 & 90.70 & 84.30 & 96.30 & 98.00 \\
& LG-MGC \cite{2024_ACMMM_1} 
& 51.6 & 77.2 & 85.7 & 66.3 & 87.7 & 93.4 & 80.3 & 96.2 & 98.4 & 92.4 & 99.2 & 99.6  \\
& CUSA \cite{Huang_Nie_Wang_Shang_2024} 
& 52.4 & 79.8 & 88.1 & 67.9 & 90.3 & 94.7 & 77.4 & 95.5 & 97.7 & 90.8 &  99.1 & 99.7 \\      
    
\hline\hline

\multirow{4}{*}{\rotatebox{90}{w/ CLIP}}
& CLIP-ViT-B/32 & 42.83 & 71.24 & 81.13 & 56.34 & 81.76 & 89.42 & 66.33 & 88.62 & 93.13 & 78.72 & 95.42 & 98.03 \\
& \cellcolor{gray!10}\hspace{1em}\textbf{+ CalibCLIP} & \cellcolor{gray!10}\textbf{43.94} & \cellcolor{gray!10}\textbf{72.35} & \cellcolor{gray!10}\textbf{82.89} & \cellcolor{gray!10}\textbf{56.89} & \cellcolor{gray!10}\textbf{82.6} & \cellcolor{gray!10}\textbf{90.02} & \cellcolor{gray!10}\textbf{67.94} & \cellcolor{gray!10}\textbf{89.6} & \cellcolor{gray!10}\textbf{94.35} & \cellcolor{gray!10}\textbf{78.73} & \cellcolor{gray!10}\textbf{95.73} & \cellcolor{gray!10}\textbf{98.02} \\
& CLIP-ViT-L/14 & 51.63 & 79.14 & 87.72 & 67.13 & 89.43& 94.75 & 76.46 & 94.69 & 97.40 & 87.32 & 99.02 & 99.65 \\
& \cellcolor{gray!10}\hspace{1em}\textbf{+ CalibCLIP} & \cellcolor{gray!10}\textbf{52.82} & \cellcolor{gray!10}\textbf{80.11} & \cellcolor{gray!10}\textbf{89.34} & \cellcolor{gray!10}\textbf{67.73} & \cellcolor{gray!10}\textbf{90.2} & \cellcolor{gray!10}\textbf{95.18} & \cellcolor{gray!10}\textbf{78.56} & \cellcolor{gray!10}\textbf{96.57} & \cellcolor{gray!10}\textbf{99.57} & \cellcolor{gray!10}\textbf{89.84} & \cellcolor{gray!10}\textbf{99.83} & \cellcolor{gray!10}\textbf{99.91} \\
\hline
\end{tabular*}
\end{table*}

\begin{table*}[ht]
\centering
\caption{Performance Comparison on CIRR and FashionIQ Datasets}
\label{tab:CIRresults}
\begin{tabular*}{\linewidth}{@{\extracolsep{\fill}} l|l|cccc|rr|rr|rr}
\hline
 & \multirow{3}{*}{\textbf{Method}} & \multicolumn{4}{c|}{\textbf{CIRR}} & \multicolumn{6}{c}{\textbf{FashionIQ}} \\  
\cline{3-12}
& & \multicolumn{4}{c|}{\textbf{Recall@K}} & \multicolumn{2}{c|}{\textbf{Dress}} & \multicolumn{2}{c|}{\textbf{Shirt}} & \multicolumn{2}{c}{\textbf{Top\&Tee}} \\
\cline{3-6} \cline{7-8} \cline{9-10} \cline{11-12}
& & k=1 & k=5 & k=10 & k=50 & R@10 & R@50 & R@10 & R@50 & R@10 & R@50 \\
\hline\hline

\multirow{6}{*}{\rotatebox[origin=c]{90}{w/o CLIP}}
& ARTEMIS \cite{2022_iclr} & 16.96 & 46.10 & 61.31 & 87.73 & 29.04 & 53.55 & 25.56 & 50.86 & 33.58 & 50.48 \\
& MCEM \cite{2024_TIP}  & 17.48 & 46.14 & 62.17 & 88.91 & 32.11 & 59.21 & 27.28 & 52.01 & 33.92 & 62.30 \\
& NEUCORE \cite{pmlr-v243-zhao24a} & 18.46 & 49.40 & 63.57 & 89.35 & -- & -- & -- & -- & -- & -- \\
& NSFSE \cite{2024_TMM} & 20.70 & 52.50 & 67.96 & 90.74 & 31.12 & 55.73 & 24.58 & 45.85 & 31.93 & 58.37 \\
& CAFF \cite{Wan_2024_CVPR} & -- & -- & -- & -- & 35.74 & 59.85 & 35.80 & 61.94 & 38.51 & 68.34 \\
& SPIRIT \cite{2024_tomm} & 40.23 & 75.10 & 84.16 & 96.88 & 39.86 & 64.30 & 44.11 & 65.60 & 47.68 & 71.70 \\

\hline\hline

\multirow{7}{*}{\rotatebox[origin=c]{90}{w/ CLIP}}
& CLIP-ProbCR \cite{CLIP-ProbCR} & 23.32 & 54.36 & 68.64 & 93.05 & 30.71 & 56.55 & 28.41 & 52.04 & 35.03 & 61.11 \\
& CaLa-CLIP4Cir\cite{2024_SIGIR_ACM} & 35.37 & 68.89 & 80.07 & 95.86 & 32.96 & 56.82 & 39.20 & 60.13 & 39.16 & 63.83 \\
& CLIP-CD \cite{2024_CLIP-D} & 37.68 & 69.62 & 81.44 & 93.74 & 37.68 & 62.62 & 42.44 & 63.74 & 45.33 & 67.72 \\
& CLIP4CIR \cite{Baldrati_2022_CVPR} & 38.53 & 69.98 & 81.86 & 95.93 & 33.81 & 59.40 & 39.99 & 60.45 & 41.41 & 65.37 \\
& SSN\cite{Yang_Liu_Zhang_Luo_Wang_Zhang_2024} & 43.91 & 77.25 & 86.48 & 97.45 & 34.36 & 60.78 & 38.13 & 61.83 & 44.26 & 69.05 \\

\cline{2-12}
& CLIP4CIR2 \cite{CLIP4cir2} & 42.05 & 76.13 & 86.51 & 97.49 & 37.67 & 63.16 & 39.87 & 60.84 & 44.88 & 68.59 \\
& \cellcolor{gray!10}\hspace{1em}\textbf{+ CalibCLIP} & \cellcolor{gray!10}\textbf{45.50} & \cellcolor{gray!10}\textbf{78.02} & \cellcolor{gray!10}\textbf{87.63} & \cellcolor{gray!10}\textbf{98.13} & \cellcolor{gray!10}\textbf{41.92} & \cellcolor{gray!10}\textbf{62.51} & \cellcolor{gray!10}\textbf{39.90} & \cellcolor{gray!10}\textbf{64.72} & \cellcolor{gray!10}\textbf{46.66} & \cellcolor{gray!10}\textbf{70.76} \\
\hline
\end{tabular*}
\end{table*}

Then, we establish a high-recall candidate subspace $\mathcal{C}_k$ by selecting top-$k$ matches via $\text{sim}(\mathbf{t}_{\text{eot}}, \mathbf{v}_{\text{cls}})$. Within $\mathcal{C}_k$, we compute fine-grained similarity:  
\begin{equation}
\text{sim}_{\text{disc}}(\mathbf{\hat{t}}_{\text{r}}, \mathbf{v}_{\text{cls}}^{(i)}) = \frac{\langle \mathbf{\hat{t}}_{\text{r}}, \mathbf{v}_{\text{cls}}^{(i)} \rangle}{\|\mathbf{\hat{t}}_{\text{r}}\| \cdot \|\mathbf{v}_{\text{cls}}^{(i)}\|}
\end{equation}  
where $\mathbf{v}_{\text{cls}}^{(i)}$ denotes the visual feature of the $i$-th candidate in $\mathcal{C}_k$. The final ranking score for cross-modal retrieval dynamically fuses global and fine-grained similarities with additional discriminative cues:  
\begin{equation}
\text{score} = \lambda \cdot \text{sim}(\mathbf{t}_{\text{eot}}, \mathbf{v}_{\text{cls}}) + (1-\lambda) \cdot \text{sim}_{\text{disc}}(\mathbf{\hat{t}}_{\text{r}}, \mathbf{v}_{\text{cls}}^{(i)})
\end{equation}  
where \( \lambda \in [0,1] \) is a hyper-parameter that balances the effect brought by the discriminative similarity $\text{sim}_{\text{disc}}$.

\section{Experiment}

\subsection{Implementation Details}
\label{sec:Implementation_Details}
In this study, we employed CLIP-ViT-B/32, B/16, and L/14 as baseline models, conducting all experiments on eight NVIDIA 4090 GPUs. The models were trained using the AdamW optimizer with a learning rate linearly decayed from $1 \times 10^{-4}$ to $1 \times 10^{-5}$. 

\subsection{Benchmark and Metrics}
\label{sec:subsection}
We conduct comprehensive evaluation spanning: 1) fine-grained distinction on TBPR benchmarks (CUHK-PEDES \cite{Li_2017_CVPR}, ICFG-PEDES \cite{DBLP_1}, and RSTPReid \cite{DSSL}; 2) global semantic alignment using Flickr-30K \cite{Flick30k} and MSCOCO \cite{MSCOCO} for TIR; and 3) compositional reasoning through CIR benchmarks (FashionIQ \cite{Wu_2021_CVPR} and CIRR \cite{Liu_2021_ICCV}), employing standard Recall@K metrics (K=1,5,10,50).

\begin{figure*}[t]
\centering
\includegraphics[width=.99\textwidth]{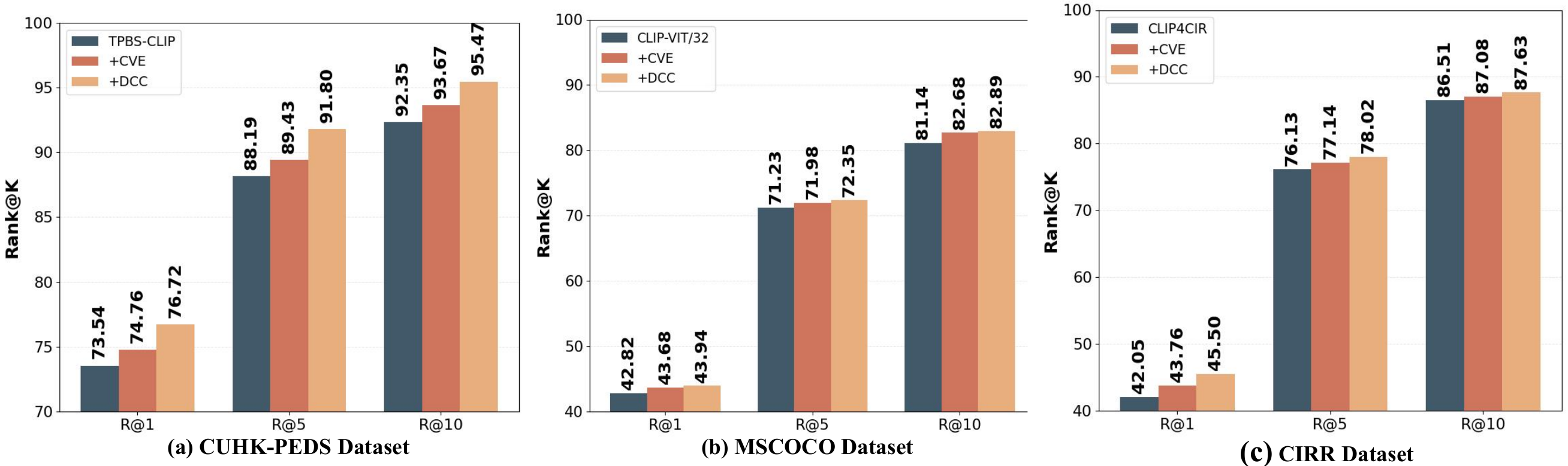} 
\caption{Ablation study of each component of our method on representative datasets for three language-driven retrieval tasks.} 
\label{fig:Ablation}
\end{figure*}

\begin{figure*}[t]
\centering
\includegraphics[width=.99\textwidth]{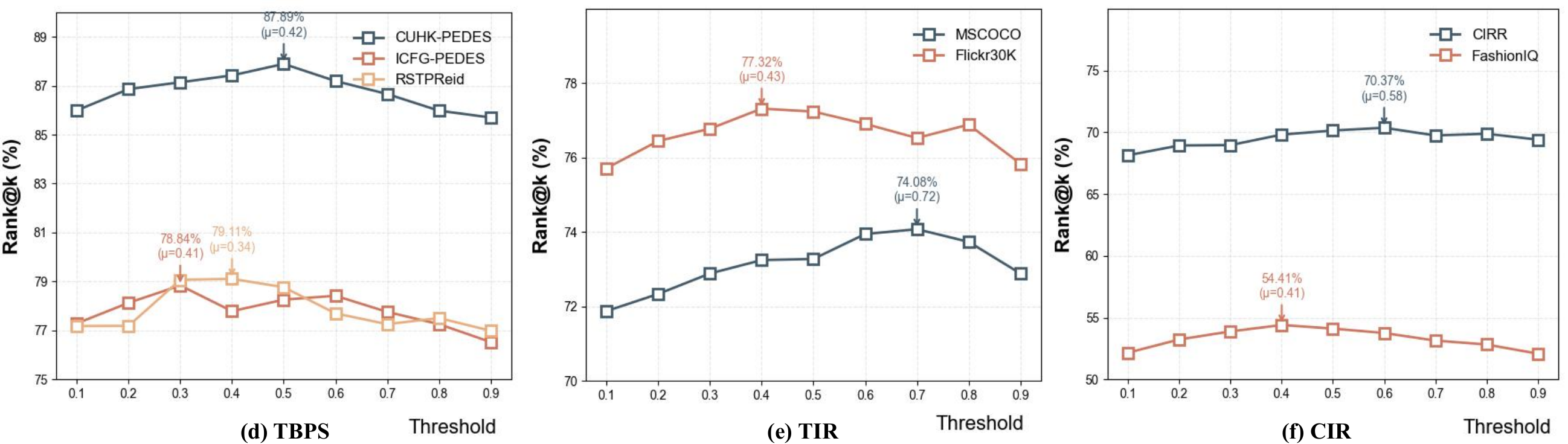} 
\caption{Visual decoupling threshold analysis on seven bases of TBPR Task (d), TIR Task (e), CIR Task (f).} 
\label{fig:Ablation_2}
\end{figure*}

\subsection{Benchmark Results}
\label{sec:subsubsection}
\mypara{Results on Text-based Person Retrieval.}
Table~\ref{tab:TBPRresults}  summarizes comprehensive evaluation results across three fine-grained retrieval benchmarks. On CUHK-PEDES, CalibCLIP achieves Rank@K improvements of 3.28\% over the CLIP-ViT-B/16 baseline in zero-shot adaptation without additional training data. Furthermore, when transferred to domain-specifically trained TPBS-CLIP, CalibCLIP achieves a notable increase in Rank-1 accuracy to 76.72\%, establishing new state-of-the-art performance. These improvements validate CalibCLIP's effectiveness in fine-grained perception and cross-modal correlation.

\mypara{Results on Text to Image Retrieval.}
Table~\ref{tab:TIRresults} summarizes comprehensive evaluation results across standard TIR benchmarks. For Flickr30K retrieval tasks, we observe consistent performance gains with a 1.63\% average Rank@k improvement. When extended to the more challenging MSCOCO dataset, the model maintains robust performance with a 1.25\% average Rank@k improvement. Performance gains become more pronounced with the scaled CLIP-ViT-L/14 architecture, especially for detail-oriented retrieval. These results confirms CalibCLIP's effectiveness in eliminating cross-modal noise.

\mypara{Results on Compose Image Retrieval.}
Table~\ref{tab:CIRresults} presents comparative results on mainstream CIR benchmarks: CIRR and FashionIQ. Despite compositional retrieval complexity, CalibCLIP achieves consistent gains across tasks through our enhanced cross-modal matching framework without architectural modifications. On the CIRR benchmark, CalibCLIP achieves 2.07\% relative improvement in Rank@k over state-of-the-art VLM adaptation methods. For FashionIQ's multi-attribute retrieval, CalibCLIP obtains 1.77\% average relative gains in Rank@10/50 across Dress, Shirt, and Top\&Tee subcategories. These results substantiate CalibCLIP's robustness and generalizability in addressing fundamental representation limitations of VLMs.

\subsection{Ablation Study}
\label{sec:Ablation}

\subsubsection{Component Efficacy Evaluation.}
\label{sec:Component_Efficacy_Evaluation}
To validate component effectiveness, we conduct ablation studies on representative datasets across three subtasks. On CUHK-PEDES, the CVE module improves the average Rank@k by 1.49\% by suppressing outlier tokens in low-information regions. Textual dominance mitigation through the DCC module provides an additional 2.37\% gain by redistributing semantic attention. MSCOCO's scene-level captions restrict fine-grained disentanglement. This results in relatively smaller contributions from dominance attenuation (0.83\% Rank@k improvement) compared to other tasks. For image retrieval with noisy queries demonstrates a stronger impact (3.12\% Rank@k boost) due to prevalent outlier patterns. Both modules enhance performance on complex CIRR queries, achieving 1.61\% and 1.51\% respective improvements through dual-path refinement.

\subsubsection{Visual Feature Decoupling Threshold.}
\label{sec:Visual_Feature_Decoupling_Threshold}
We first evaluated the effect of the adaptive thresholding strategy introduced in Section~\ref{sec:contrastive_visual_enhancer} across seven benchmark datasets. As shown in Figure~\ref{fig:Ablation_2} (d-f), the configuration using the mean ($\mu$) of cosine similarity consistently yield-ed the highest performance among the different formulations tested. This result suggests that this formulation effectively captures the distributional characteristics of attention values, enabling more reliable thresholding across diverse data scenarios.

\section{Conlusion}
In this paper, we have identified a crucial limitation in current VLMs for text-driven image retrieval: the unsupervised aggregation of global tokens disproportionately amplifies low information tokens while diminishing discriminative features. To tackle this issue, we introduce CalibCLIP, a training-free framework that incorporates dual calibration mechanisms for both visual and textual spaces. Our approach dynamically suppresses spatial outliers in visual features through contrastive localization and enhances text representations by disentangling general and discriminative semantic concepts. 

\mypara{Acknowledgement.}
This work was supported by the Guangdong Basic and Applied Basic Research Foundation (2025A15150115-46), the Shenzhen Science and Technology Innovation Program (JCYJ20240813105901003, KJZD20240903102901003), and the Science and Technology Project of Shenzhen (GXWD20220811170603-002).

\balance

\end{document}